\documentclass[conference]{IEEEtran}
\IEEEoverridecommandlockouts
\usepackage{silence}
\usepackage{caption}
\usepackage{cite}
\usepackage[table]{xcolor}
\definecolor{lightgreen}{RGB}{220,255,220}
\usepackage{placeins}
\usepackage[font=footnotesize]{caption}
\usepackage{graphicx}
\usepackage{textcomp}
\usepackage{xcolor}
\usepackage{float}
\usepackage{todonotes}
\usepackage{booktabs}
\usepackage{caption}
\usepackage{algorithm}
\usepackage{algpseudocode}
\usepackage{amsmath}
\usepackage{multirow}
\usepackage[colorlinks=true, linkcolor=blue, citecolor=blue, urlcolor=blue]{hyperref}
\usepackage{listings}  
\usepackage{xcolor}    

\lstdefinestyle{codeStyle}{
    basicstyle=\ttfamily\small,
    breaklines=true,        
    frame=single,           
    backgroundcolor=\color{gray!10},
    postbreak=\mbox{\textcolor{red}{$\hookrightarrow$}\space},
}

\def\BibTeX{{\rm B\kern-.05em{\sc i\kern-.025em b}\kern-.08em
    T\kern-.1667em\lower.7ex\hbox{E}\kern-.125emX}}
\begin{document}

\title{Conference Paper Title*\\

\thanks{Identify applicable funding agency here. If none, delete this.}
}


\title{Structure-Aware Chunking for Tabular Data in Retrieval-Augmented Generation}

\author{
Pooja Guttal\quad
Varun Magotra\quad
Vasudeva Mahavishnu\quad
Natasha Chanto\\
Sidharth Sivaprasad\quad
Manas Gaur\\[0.5em]
{\small \texttt{\{pooja, varun, vasu, natasha, sidharth\}@altumatim.com}}\\
{\small \texttt{manas@umbc.edu}}\\[0.3em]
Altumatim, \quad University of Maryland, Baltimore County
}

\maketitle

\begin{abstract}
Tabular documents such as CSV and Excel files are widely used in enterprise data pipelines, yet existing chunking strategies for retrieval-augmented generation (RAG) are primarily designed for unstructured text and do not account for tabular structure. We propose a structure-aware tabular chunking (STC) framework that operates on row-level units by constructing a hierarchical Row Tree representation, where each row is encoded as a key-value block.

STC performs token-constrained splitting aligned with structural boundaries and applies overlap-free greedy merging to produce dense, non-overlapping chunks. This design preserves semantic relationships between fields within a row while improving token utilization and reducing fragmentation.

Across evaluations on the MAUD dataset, STC reduces chunk count by up to 40\% and 56\% compared to standard recursive and key-value–based baselines, respectively, while improving token utilization and processing efficiency. In retrieval benchmarks, STC improves MRR from 0.3576 to 0.5945 in a hybrid setting and increases Recall@1 from 0.366 to 0.754 in BM25-only retrieval.

These results demonstrate that preserving structure during chunking improves retrieval performance, highlighting the importance of structure-aware chunking for RAG over tabular data.
\end{abstract}

\section{Introduction}
\label{sec:introduction}

Document chunking is a critical design component in retrieval augmented systems, as it determines how information is segmented, represented, and retrieved. Existing chunking strategies such as fixed-size chunking \cite{lewis2020rag}, sliding window approaches \cite{beltagy2020longformer}, and content-aware segmentation \cite{hearst1997texttiling} are primarily designed for unstructured text. These methods assume linear and semantically continuous text, which limits their effectiveness for tabular data, where information is structured across hierarchical rows, columns, and interdependent field-level relationships.\cite{chunking_study_2026, lewis2020rag}.

Tabular documents such as CSV and Excel files introduce additional challenges due to their structured schema, heterogeneous cell contents, and the need to preserve row and table-level context. Naive linearization of such data often leads to truncated context, loss of relational information, and semantically incoherent chunks. Prior work shows that suboptimal chunking can significantly degrade retrieval performance, as arbitrary segmentation, fragments context and reduces retrieval accuracy compared to structure-preserving segmentation approaches. \cite{chunking_study_2026}. Similarly, fixed-size chunking in retrieval augmented generation systems can result in incomplete retrieval and reduced generation coherence due to the loss of global context \cite{rag_chunking_2025}.

Recent work in code and structured document processing, such as CAST \cite{cast2025}, demonstrates that structure-aware chunking based on Abstract Syntax Trees (ASTs) can preserve hierarchical relationships by recursively splitting large nodes and greedily merging them under size constraints. This approach produces self-contained, semantically coherent chunks that improve downstream retrieval and generation quality. However, such methods rely on syntactic tree representations derived from programming languages and specialized parsing mechanisms, which are not directly applicable to tabular data. This motivates the need for a structure-aware chunking approach tailored to tabular data, where hierarchical relationships exist.

In this work, we adapt the CAST (Chunking via Abstract Syntax
Trees) paradigm~\cite{cast2025} for tabular data by introducing a Row Tree representation that captures row and table-level structure. Building on this representation, we propose a Structure-Aware Tabular Chunking (STC) framework, where each row is encoded as a structured key-value block. This enables token-constrained splitting and greedy merging to operate on structured units while maintaining local context within each chunk. The proposed method produces dense, non-overlapping chunks and reduces fragmentation, while operating in linear time with respect to the number of rows.

Our contributions are as follows:
\begin{enumerate}
    \item We propose a structure-aware chunking method for tabular documents, inspired by CAST~\cite{cast2025}, but tailored for CSV and Excel data.
    \item We introduce a Row Tree structure and key-value (KV) representation that organizes tabular data into structured units, grouping related attributes within the same chunk and reducing fragmentation under token constraints.
    \item We adapt split and greedy merge strategies for tabular data, producing dense, non-overlapping chunks and achieving linear-time processing with respect to the number of rows.
\end{enumerate}

\begin{figure*}[t]
\centering
\includegraphics[width=\textwidth]{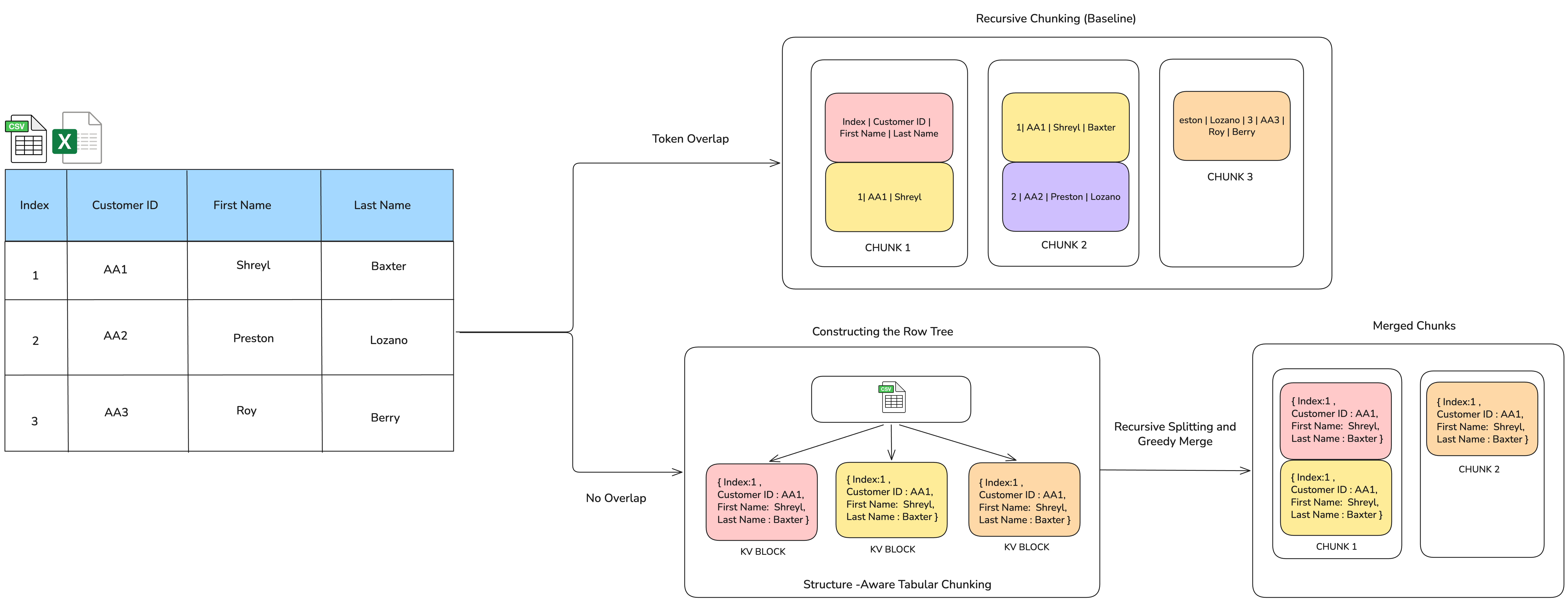}
\caption{Comparison of baseline recursive chunking versus the proposed structure-aware framework. The baseline method (top) operates on linearized text with token-level splitting and overlap, which can fragment rows across multiple chunks and introduce redundant tokens. In contrast, the proposed framework (bottom) represents each row as a key-value (KV) block within a Row Tree hierarchy. By applying token-constrained splitting and greedy merging, the method produces dense, non-overlapping chunks aligned with row-level structure.}
\label{fig:router}
\end{figure*}

\section{Related Work}

Chunking is a critical preprocessing step in retrieval RAG pipelines, where long documents are divided into smaller segments for efficient retrieval and reasoning. Common approaches rely on heuristic-based text splitting, such as fixed-size windows or separator-based methods (e.g., LangChain’s RecursiveCharacterTextSplitter), often combined with sliding-window overlap to preserve context~\cite{LangChain_2022}. However, these methods are primarily designed for unstructured text and do not account for structural boundaries, leading to fragmented chunks and redundant tokens due to overlap.

Prior work has shown that chunk size and segmentation strategies significantly impact retrieval effectiveness in RAG systems~\cite{lewis2020retrieval}. At the same time, long contexts are often underutilized by language models, motivating the need for efficient and well-structured chunking~\cite{liu2023lost}. Despite these findings, most approaches evaluate chunking indirectly through downstream task performance, rather than assessing the quality of the chunking process itself.

Several works have explored structure-aware document processing and segmentation techniques. CAST~\cite{cast2025} introduces a code-aware chunking approach using abstract syntax trees to preserve program structure during splitting. In parallel, text segmentation has been studied as a supervised learning problem, focusing on identifying coherent boundaries within documents~\cite{koshorek2018text}. However, these approaches primarily target natural language or code and do not directly address the challenges of tabular data.

A separate line of work focuses on modeling tabular data for language understanding, including TaBERT~\cite{yin2020tabert}, TAPAS~\cite{herzig2020tapas}, and TURL~\cite{deng2020turl}, which learn joint representations of text and tables. These models leverage structural information present in tabular inputs to enable reasoning over rows and columns. However, their effectiveness depends on how the input is represented, and they do not explicitly focus on how chunking strategies preserve or disrupt this structure in RAG pipelines. In contrast, our work focuses on designing chunking methods that better retain tabular structure, enabling more faithful downstream representations.

Existing chunking methods are largely agnostic to tabular structure, while structure-aware approaches focus on code or natural language documents rather than row-based data. As a result, there is limited work on chunking strategies tailored to tabular datasets. We address this gap by introducing a row-structured chunking framework that organizes tabular data into structured units, enabling efficient chunk construction, improved token utilization, and reduced fragmentation under token constraints.

\section{Methodology}

We propose a structure-aware chunking framework for tabular documents designed for downstream RAG pipelines. Instead of linearizing the input, the method organizes data at the row level, maintaining local context within each chunk. The framework supports both single-table formats (e.g., csv) and multi-sheet workbooks (e.g., xlsx).

As illustrated in Fig.~\ref{fig:router}, the pipeline consists of three stages: (i) constructing a hierarchical Row Tree representation, (ii) performing token-budget–constrained splitting, and (iii) applying greedy merging to produce the final chunks.

\subsection{Row Tree Representation}

As shown in Fig.~\ref{fig:router}, the input table is transformed into a hierarchical Row Tree representation that organizes tabular data into structured units for chunking. 

The Row Tree representation is constructed in a unified manner for tabular inputs. Each table is represented as a hierarchical structure with a root node and row-level nodes corresponding to individual records. Each row is converted into a structured key-value (KV) format, where non-empty cells are encoded as \texttt{column\_name: value} pairs.

The hierarchy depth depends on the input format: single-table inputs (e.g., csv) yield a root-to-row structure, while multi-sheet inputs (e.g., xlsx) introduce an intermediate sheet-level node between the root and row-level nodes. 

This representation organizes tabular data into consistent structural units and maintains relationships between fields within each row, enabling structure-aware processing in subsequent stages.

\subsection{Recursive Splitting}
Leveraging the Row Tree representation, splitting is formulated as a top-down hierarchical traversal under a maximum token constraint. Each node in the tree represents a structured unit (e.g., sheet or row).

During traversal, each node is evaluated against the max token constraint. If the node satisfies the constraint, it is retained as a leaf unit. Otherwise, the node is decomposed into its child units, and the process is applied recursively until all resulting units satisfy the token limit.

In cases where a leaf node (e.g., a row) itself exceeds the token budget, a secondary splitting procedure is applied. As described in Section~\ref{sec:emergency_split}, this procedure performs field-aligned splitting at key-value boundaries, ensuring that oversized rows are divided into smaller units that remain consistent with the underlying tabular structure.

By operating on structured nodes rather than linearized text, this approach aligns splitting with the inherent organization of the data. As shown in Fig.~\ref{fig:router}, this avoids the fragmentation that can arise when splitting is applied directly to raw text. The output of this stage is a set of leaf-level units that satisfy the token constraint and serve as input to the merging stage.

\subsection{Emergency Splitting}
\label{sec:emergency_split}

In cases where a row-level unit exceeds the maximum token constraint and cannot be further decomposed within the Row Tree hierarchy, a fallback splitting procedure is applied.

The row is treated as an ordered sequence of key-value (KV) pairs, and splitting is performed at field boundaries. KV pairs are accumulated sequentially until adding the next pair would exceed the token limit, at which point a new fragment is created. This process continues until the entire row is partitioned into fragments that satisfy the token constraint.

By splitting at KV boundaries, each fragment contains complete fields rather than partial text segments. The resulting fragments are non-overlapping and are used as leaf-level units for the subsequent merging stage.

\subsection{Greedy Merging}
To reduce the total number of chunks and improve token utilization, a greedy merging strategy is applied to adjacent leaf nodes (Fig.~\ref{fig:router}, right). Leaf nodes are merged within the same parent node in the Row Tree (e.g., same table or sheet), ensuring that merging operates within consistent structural boundaries.

Within each parent node, leaf nodes are accumulated sequentially until adding the next node would exceed the token limit, at which point a new chunk is started. This results in dense, non-overlapping chunks that make effective use of the available token capacity.

The resulting chunks are non-overlapping and constructed under the token constraint, eliminating the need for overlap-based redundancy. Each chunk consists of multiple row-level nodes in key-value (KV) form and is generated within the same parent node in the Row Tree, ensuring consistent structure across chunks. These chunks serve as structured inputs for downstream retrieval and generation tasks.

\begin{algorithm}[H]
\caption{STC Chunking for Tabular Documents}
\label{alg:rsm}
\begin{algorithmic}[1]
\Require Tabular document $D$, token budget $B$
\Ensure List of chunks $C$
\State $H \gets$ column headers of $D$
\State $L \gets \emptyset$
\For{each row $r_i$ in $D$}
    \State $n_i \gets$ key-value block of $r_i$ with context $H$
    \If{TokenCount$(n_i) \leq B$}
        \State $L \gets L \cup \{n_i\}$
    \Else
        \State $L \gets L \cup \text{SplitOnKeyValue}(n_i)$
    \EndIf
\EndFor
\State $C \gets \emptyset$
\For{each context group $G$ in $L$}
    \State greedily batch leaves of $G$ into chunks within $B$
    \State append chunks to $C$
\EndFor
\State \Return $C$
\end{algorithmic}
\end{algorithm}

\section{Dataset}
\subsection{MAUD (Merger Agreement Understanding Dataset)}

We evaluate our approach on the Merger Agreement Understanding Dataset (MAUD)~\cite{maud}, a benchmark for legal reading comprehension derived from merger and acquisition (M\&A) contracts sourced from the SEC EDGAR system~\cite{sec_edgar}. MAUD consists of expert-annotated question-answer pairs grounded in real-world legal agreements and is widely used in legal NLP research. Each MAUD data instance represents a structured record consisting of an extracted deal point text, a deal point question, and one or more predefined answers. The deal point text corresponds to a contract clause extracted from a merger agreement, while the question and answer fields capture the legal interpretation associated with that clause.

In addition to the clause text, each record includes structured attributes such as deal point category, deal point type, question, answer, and contract identifier. Together, these fields form a semi-structured representation, where the \textit{text} field contains unstructured legal language and the remaining fields provide structured information.

This structure allows each record to be interpreted as a collection of key-value (KV) pairs, where each attribute (e.g., \textit{text}, \textit{question}, \textit{answer}) serves as a key associated with its value. Such a representation aligns naturally with our approach, which models tabular data as key-value blocks within a Row Tree to preserve relationships between textual content and associated attributes during chunking.

MAUD is divided into train, validation, and test splits, as summarized in Table~\ref{tab:maud_splits}. We evaluate our approach across these splits to analyze chunking behavior over inputs of varying sizes.

\begin{table}[h]
\centering
\small
\begin{tabular}{lr}
\toprule
\textbf{Split} & \textbf{Rows} \\
\midrule
Train      & 25,827 \\
Validation &  6,753 \\
Test       &  6,651 \\
\midrule
Total      & 39,231 \\
\bottomrule
\end{tabular}
\caption{MAUD dataset statistics.}
\label{tab:maud_splits}
\end{table}



\section{Results and Analysis}
\label{sec:results}

\subsection{Chunking Evaluation Setup}
All methods are evaluated under a target budget of 512 tokens per chunk, reflecting a practical balance between context coverage and computational efficiency in RAG pipelines. Token limits are enforced using a consistent approximation across all methods, as chunk size directly affects retrieval effectiveness while excessively long contexts are often underutilized by language
models~\cite{lewis2020retrieval, liu2023lost}.

We evaluate our proposed framework against two comparative baselines. The standard Recursive approach serves as our primary baseline, utilizing LangChain’s RecursiveCharacterTextSplitter with a 100-token sliding-window overlap. To isolate the impact of our structural representation, we introduce a KV + Recursive ablation study that applies the identical recursive splitting logic to data pre-formatted into key-value pairs. Finally, our Proposed method leverages the Row Tree hierarchy alongside token-budget–constrained splitting and overlap-free greedy merging.

Performance is measured using three metrics: (i) token statistics and chunk count, including average, minimum, and maximum tokens per chunk; (ii) token utilization, defined as the ratio of average chunk size to the 512-token limit; and (iii) processing time.

 These metrics directly assess how effectively each method utilizes the token budget and organizes structured data, which are key factors for downstream performance.

\medskip
\paragraph{Chunk Statistics and Efficiency}

As shown in Table~\ref{tab:maud_chunking}, the proposed framework (RSM) achieves higher token efficiency across all data splits. By eliminating overlap and applying greedy merging, RSM produces a more compact and information-dense chunk distribution.

The proposed approach reduces the total number of chunks by approximately 40\% relative to the Recursive baseline and by over 56\% compared to KV + Recursive. In addition, it achieves higher average token utilization (approximately 399--402 tokens per chunk), indicating more effective use of the available token capacity.

The maximum token count remains bounded by the 512-token constraint across all methods. Overall, these results show that the proposed approach produces fewer, denser chunks while maintaining consistency with the token limit.

\medskip


\paragraph{Processing Speed}

The proposed framework achieves consistently lower processing speed across all splits (Table~\ref{tab:maud_chunking}). These gains are primarily driven by the use of structure-aware operations and the elimination of overlap.

Unlike baseline methods that rely on sliding-window overlap and operate on expanded text representations, the proposed approach processes structured row-level nodes directly, avoiding redundant computation across overlapping regions. In addition, greedy merging reduces the number of generated chunks, lowering the overall processing workload.

As a result, the framework achieves consistent speedups across all splits, with improvements becoming more pronounced as input size increases.

\begin{table}[t]
\centering
\scriptsize
\begin{tabular}{llrrrrr}
\toprule
\textbf{Split} & \textbf{Strategy} & \textbf{Chunks} & \textbf{Avg} & \textbf{Min} & \textbf{Max} & \textbf{Speed(ms)} \\
\midrule
\multirow{3}{*}{Train}
& Recursive        & 61,923 & 326 & 4 & 750 & 10,886.6 \\
& KV+Rec           & 87,854 & 238 & 3 & 819 & 12,218.2 \\
& \textbf{RSM}     & \textbf{42,593} & \textbf{400} & \textbf{10} & \textbf{512} & \textbf{2,498.4} \\
\midrule
\multirow{3}{*}{Test}
& Recursive        & 15,596 & 329 & 4 & 723 & 2,724.9 \\
& KV+Rec           & 22,153 & 240 & 3 & 761 & 2,751.5 \\
& \textbf{RSM}     & \textbf{10,811} & \textbf{399} & \textbf{10} & \textbf{512} & \textbf{518.7} \\
\midrule
\multirow{3}{*}{Val}
& Recursive        & 16,121 & 327 & 4 & 718 & 3,191.2 \\
& KV+Rec           & 22,888 & 239 & 3 & 742 & 2,847.4 \\
& \textbf{RSM}     & \textbf{11,057} & \textbf{402} & \textbf{10} & \textbf{512} & \textbf{529.3} \\
\bottomrule
\end{tabular}
\caption{Chunking statistics across MAUD splits (max 512 tokens).}
\label{tab:maud_chunking}
\end{table}




\subsection{Retrieval Evaluation Setup}
We evaluate all chunking strategies on the MAUD dataset under a controlled retrieval benchmark, where each method is assessed under identical indexing and retrieval conditions. The index is constructed using chunks generated from the training split. We randomly sample 1,000 records from the training data and form queries by concatenating the legal question with the corresponding contract name.

A retrieved chunk is considered relevant only if it contains both the contract name and the target question label. This strict AND condition ensures that retrieval is anchored to both the correct document and the specific legal concept. Since relevance is determined through heuristic string matching, this setup provides a controlled comparison of chunking strategies rather than an absolute measure of end-to-end retrieval performance.

We evaluate retrieval under two complementary settings. The first is a hybrid pipeline combining dense and sparse retrieval. A bi-encoder 
(\texttt{multi-qa-MiniLM-L6-cos-v1}) retrieves the top-20 candidates based on semantic similarity, while BM25 retrieves the top-5 candidates based on lexical matching. The union of these candidates is then reranked using a cross-encoder (\texttt{ms-marco-MiniLM-L6-v2}), which performs fine-grained relevance scoring.

\begin{table}[h]
\centering
\begin{tabular}{lcccc}
\toprule
\textbf{Strategy} & \textbf{R@1} & \textbf{R@3} & \textbf{R@5} & \textbf{MRR} \\
\midrule
Recursive       & 0.3470 & 0.3660 & 0.3720 & 0.3576 \\
KV + Recursive  & 0.3200 & 0.3390 & 0.3390 & 0.3296 \\
\textbf{STC}    & \textbf{0.5390} & \textbf{0.6200} & \textbf{0.6550} & \textbf{0.5945} \\
\bottomrule
\end{tabular}
\caption{Retrieval performance under the hybrid setting combining dense (bi-encoder) and sparse (BM25) retrieval with cross-encoder reranking. STC consistently outperforms both baselines across Recall@k and MRR while using fewer chunks, indicating improved chunk quality and a more efficient index.}
\label{tab:overall}
\end{table}
The second setting uses BM25 alone as a lexical baseline, allowing us to isolate the impact of chunk structure on sparse retrieval without the influence of learned embeddings.

Performance is evaluated using Recall@\{1, 3, 5\} and Mean Reciprocal Rank (MRR), capturing both the ability to retrieve relevant chunks within the top-$k$ results and the ranking quality of the first relevant hit.

\paragraph{Retrieval Performance Analysis}
Tables~\ref{tab:overall} and~\ref{tab:bm25} present the retrieval performance of the evaluated chunking strategies under hybrid and BM25-only settings, respectively.

In the hybrid setting (Table~\ref{tab:overall}), STC outperforms both baselines across all metrics. In particular, it achieves a substantially higher MRR compared to the Recursive baseline, indicating that preserving structural information during chunking improves both ranking quality and semantic matching.

The effect of the STC strategy is even more pronounced in the BM25-only setting (Table~\ref{tab:bm25}). It significantly improves Recall@1 over the Recursive baseline, suggesting that the resulting chunks are more lexically coherent. This is because STC preserves row-level structure and performs splitting at key-value boundaries only when required to satisfy the token constraint, ensuring that related fields remain within the same chunk and reducing boundary fragmentation, an important factor for sparse retrieval.

In contrast, the KV + Recursive ablation consistently underperforms the Recursive baseline in both settings. Although the data is converted into a key-value format, the use of standard text-based recursive splitting with overlap ignores structural boundaries. As a result, key-value pairs and row-level context are frequently fragmented across chunks, weakening local coherence and negatively affecting both dense and sparse retrieval performance.

\section{Conclusion and Limitations}

This work demonstrates that effective chunking for tabular data requires preserving structural boundaries rather than relying on token-based text splitting. The proposed STC framework operates on row-level units, maintaining record integrity and applying key-value partitioning only when required by token constraints. This enables the creation of dense, non-overlapping chunks that retain meaningful relationships between fields.

Empirical results on the MAUD dataset show that these structural properties improve both efficiency and retrieval performance, yielding fewer, more informative chunks and higher performance in both dense and sparse retrieval settings.

\begin{table}[h]
\centering
\begin{tabular}{lcccc}
\toprule
\textbf{Strategy} & \textbf{R@1} & \textbf{R@3} & \textbf{R@5} & \textbf{MRR} \\
\midrule
Recursive       & 0.3660 & 0.3760 & 0.3790 & 0.3720 \\
KV + Recursive  & 0.3260 & 0.3340 & 0.3370 & 0.3305 \\
\textbf{STC} & \textbf{0.7540} & \textbf{0.7890} & \textbf{0.8030} & \textbf{0.7757} \\
\bottomrule
\end{tabular}
\caption{Retrieval performance under the BM25-only setting. STC achieves substantially higher Recall@k and MRR, demonstrating that STC produces more lexically coherent chunks that benefit sparse retrieval.}
\label{tab:bm25}
\end{table}

Although the evaluation is conducted on a legal dataset, the underlying approach is not domain-specific. STC is designed for tabular data and can be applied to other structured sources such as spreadsheets, logs, and database exports, where preserving relationships between fields is essential for downstream tasks.

The current evaluation is limited to a fixed token budget and a controlled retrieval setup using heuristic relevance matching. Future work will extend this analysis to varying token budgets, more diverse tabular datasets, and fully end-to-end RAG pipelines to assess the impact on generation quality and reasoning tasks.

Overall, preserving structure in chunking improves retrieval over tabular data and guides handling structured inputs in RAG systems.


\bibliographystyle{IEEEtran}
\nocite{*}
\bibliography{references}

\newpage

\end{document}